\documentclass[review]{elsarticle}

\usepackage{multirow}
\usepackage{xcolor, float, caption}
\usepackage{booktabs}
\usepackage{amsmath}
\usepackage{hyperref}
\usepackage{graphicx,subcaption}
\usepackage{epstopdf}
\usepackage{geometry}
\usepackage{tabularx}
\usepackage{threeparttable}
\begin{document}

\begin{frontmatter}
\title{A Novel Automation-Assisted Cervical Cancer Reading Method Based on Convolutional Neural Network}
\author{Yao Xiang}
\ead{yao.xiang@cus.edu.cn}
\author{Wanxin Sun}
\ead{sunwanxin@cus.edu.cn}
\author{Changli Pan}
\ead{changlip@csu.edu.cn}
\author{Meng Yan}
\ead{bryant@cus.edu.cn}
\author{Zhihua Yin}
\ead{yzhuajd@cus.edu.cn}
\author{Yixiong Liang\corref{mycorrespondingauthor}}
\cortext[mycorrespondingauthor]{Corresponding author}
\ead{yxliang@cus.edu.cn}
\address{School of Computer Science and Engineering, Central South University, Changsha 410083, China}

\begin{abstract}
Cervical cytology screening using Pap smear or liquid-based cytology is one of the most widely followed and accepted method. Automation-assisted screening based on cervical cytology has become a necessity due to the manual screening method operated by a visual analysis for cervical cell specimen under the microscope of the glass slide is usually labor-intensive and time-consuming. While automation-assisted reading system can improve efficiency, their performance often relies on the success of accurate cell segmentation and hand-craft feature extraction. This paper presents an efficient and totally segmentation-free method for automated cervical cell screening that utilizes modern object detector to directly detect cervical cells or clumps, without the design of specific hand-crafted feature. Specifically, we use the state-of-the-art CNN-based object detection methods, YOLOv3, as our baseline model. In order to improve the classification performance of hard examples which are four highly similar categories, we cascade an additional task-specific classifier. We also investigate the presence of unreliable annotations and coped with them by smoothing the distribution of noisy labels. We comprehensively evaluate our methods on our test set which is consisted of 1,014 annotated cervical cell images with size of 4000$\times$3000 and complex cellular situation corresponding to 10 categories. Our model achieves 97.5\% sensitivity (\emph{Sens}) and 67.8\% specificity (\emph{Spec}) on cervical cell image-level screening. Moreover, we obtain a best mean Average Precision (mAP) of 63.4\% on cervical cell-level diagnosis, and improve the Average Precision (AP) of hard examples which are the most valuable but most difficult to distinguish. Our automation-assisted cervical cell reading system not only achieves cervical cell image-level classification but also provides more detailed location and category reference information of abnormal cells. The results indicate feasible performance of our method, together with the efficiency and robustness, providing a new idea for future development of computer-assisted reading systems in clinical cervical screening.
\end{abstract}

\begin{keyword}
Cervical cancer screening \sep Papanicolaou test \sep cervical cytology \sep convolutional neural network \sep the Bethesda system
\end{keyword}

\end{frontmatter}
%\linenumbers

% 1 Introduction
\section{Introduction}\label{section1}
Cervical cancer is one of the most common causes of cancer death for women worldwide and is most frequently in developing countries \cite{william2018review}. Papanicolaou test (abbreviated as Pap test) or cervical cytology is now a mainstay cervical cancer screening method to detect potentially pre-cancerous and cancerous process in the cervix, which has demonstrated reduction in cervical cancer incidence and mortality in developed countries\cite{mishra2011overview, bora2017automated}. Such method is performed by a visual examination of cytopathological analysis under the microscope of the glass slide and finally giving a diagnosis report according to the descriptive diagnosis method of the Bethesda system (TBS) \cite{nayar2015bethesda}. However, manual analysis of microscope images is time-consuming, labor-intensive and error-prone as a handful of abnormal cells among millions of cells within a single slide has to be identified by a trained professional \cite{william2019pap}.

\begin{figure}[t]
\centering
\begin{subfigure}[b]{1.0\textwidth}
\includegraphics[width=1.0\textwidth]{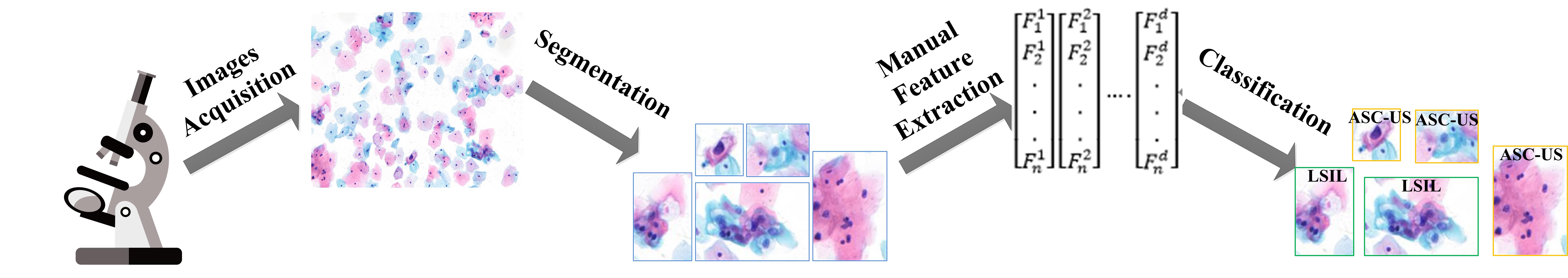}
\caption{the traditional multi-stage pipeline}
\label{s1}
\includegraphics[width=0.95\textwidth]{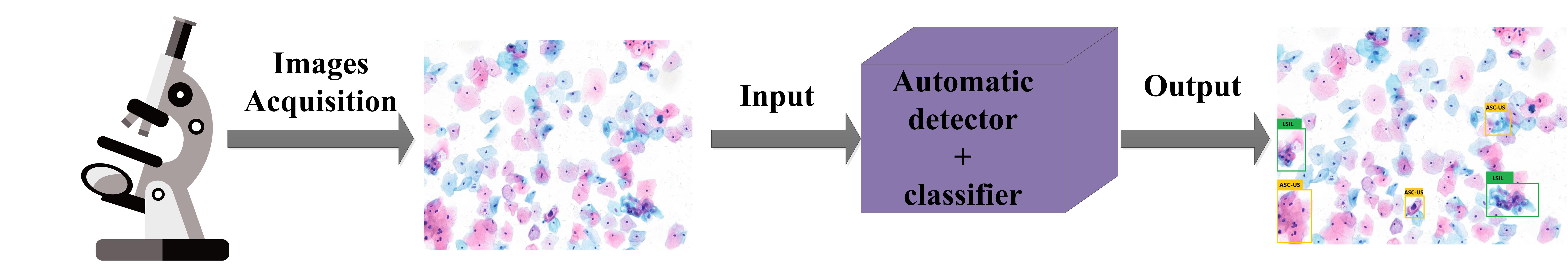}
\caption{the proposed end-to-end pipeline}
\label{s2}
\end{subfigure}
\caption{\label{figure1}The pipelines for cervical recognition. (a) the traditional multi-stage pipeline, (b)the proposed end-to-end pipeline.}
\end{figure}

Therefore, automation-assisted screening based on cervical cytology has become a necessity. Since the first system was developed in 1950s, extensive research has attempted to exploit automation-assisted reading systems based on automatic image analysis techniques(e.g.\cite{birdsong1996automated, koss1994evaluation, wilbur1998autopap}) which led to a couple of commercial systems emerged, such as the BD FocalPoint Slide Profiler \cite{wilbur2009becton} and ThinPrep \cite{biscotti2005assisted} which received approval from the US Food and Drug Administration(FDA). While automation-assisted reading systems can increase productivity by reducing the time needed to read slides, their current performance and costs are not recommended for application in primary cervical screening \cite{kitchener2011automation, bengtsson2014screening}. To this end, lots of automation-assisted methods based on cervical cell image analysis have been proposed \cite{william2019pap, gencctav2012unsupervised, zhang2014automation, sarwar2015hybrid, sharma2016various}.
Most of them follow the multi-stage pipeline, i.e., first identifying the candidate regions based on segmentation, then extracting hand-crafted features based on the characteristics of nuclei and cytoplasm for classification, as shown in Fig.~\ref{figure1} (a).
% figure 1

\begin{figure}[htbp]
\centering
\begin{subfigure}[b]{0.4\textwidth}
\includegraphics[width=\textwidth]{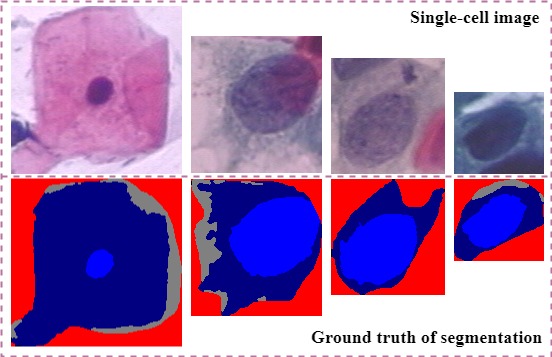}
\caption{}
\label{figure2_1}
\end{subfigure}
\begin{subfigure}[b]{0.45\textwidth}
\includegraphics[width=\textwidth]{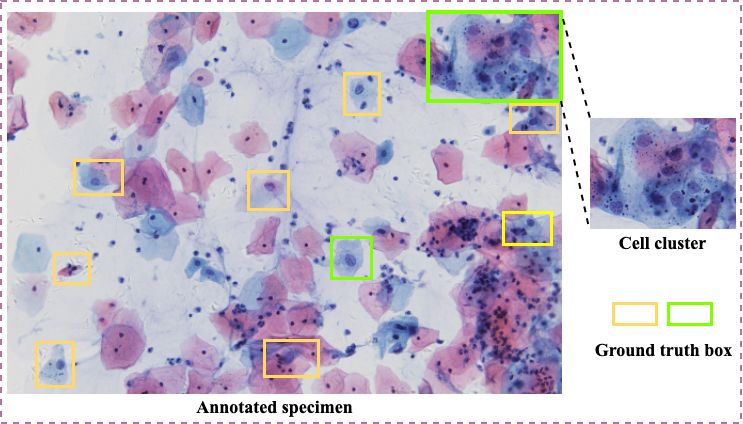}
\caption{}
\label{figure2_2}
\end{subfigure}
\caption{\label{figure2}Example images and ground truth of single-cell and multi-cell image from the (a) Herlev and (b) our dataset. Each single-cell image from Herlve has an associated ground truth of nucleus and cytoplasm regions for segmentation, and the average image size of them is 156$\times$140 pixels. Multi-cell specimen from our dataset has many ground truth boxes of isolated cells and cell clusters for detection, and each specimen has 4000$\times$3000 pixels.}
\end{figure}

While most of these studies have achieved available performance whether in cell segmentation or cell classification, there still have some challenges to use them in clinical automation-assisted reading. First, the current automation-assisted reading approaches has not been sufficiently cost-effective to promote to the cervical cell screening in clinical due to the tedious image patches preprocessing and screening process for cyto-technicians and doctors. The majority of existing research has done on the Herlve dataset \cite{jantzen2005pap} which only contains single-cell images with a size of 200$\times$100 pixels approximately and was produced carefully by trained professionals. As shown in Fig.~\ref{figure2}, images from Herlve dataset are all clear with no overlapping and impurity. In fact, the slide image with around 2,160 million pixels obtained by the Whole Slide Imaging (WSI) technology \cite{Fertig2017Whole} has complex cellular situation, such as cell overlapping, noise and impurity. Thus,one original cervical cell slide should be cropped into a huge number of single-cell image patches by using sliding windows or region proposal generation methods based on low-level image features, which lead to low efficiency. Second, it is difficult to make segmentation of the cytoplasm and nuclei absolutely due to the high degree of cell overlapping, the poor contrast of the cell cytoplasm and the presence of mucus, noise and impurity. Third, it is worth considering that whether the hand-crafted features can represent complex identification information or not since the richer semantic information sensitive to recognition may actually exist in hidden upper-level features of cervical cell images\cite{zhang2017deeppap}. In addition, medical images are too complex and variable to get a perfect annotation as ground truth, which leads to noisy label unavoidably. However, the previous studies usually deal with cervical cell segmentation and classification without considering the existence of noisy labels.
% figure 2

To cope with these problems, we propose to utilize CNN-based object detection to automatically extract and learn task-specific features, and achieve the cervical cells recognition efficiently on multi-cell images with cell overlapping and clusters, as showing in Fig.~\ref{figure1} (b). Our method directly operates on multi-cell image with size of 4000$\times$3000 automatically extract more complex discriminative features and can obtain a image-level classification results. Moreover, it not only achieves cervical cell image-level classification but also provides more detailed location and category reference information of abnormal cells. In detail, we exploit YOLOv3 \cite{redmon2018yolov3} as our cervical cell object detection baseline model due to the efficiency, accuracy and flexibility. In order to improve the classification performance of hard examples which are four highly similar categories, we cascade a further task-specifical classifier. Furthermore, we weaken the influence of noisy labels by smoothing their distribution.

Our contributions are summarized as follows, 1) Unlike the previous method, we treat the cervical cell recognition as object detection which automatically detect cervical cells directly on multi-cell images. It is more efficient as we can extract features automatically without manual intervention and careful design for all stages. Our method not only achieves cervical cell image-level classification but also provides more detailed location and category information of abnormal cells simultaneously. 2) We propose a simple and effective scheme, cascade a further task-specific classifier to improve the performance of hard example recognition. 3) We investigate the existence of noisy labels on cervical cell dataset and propose an approach to weaken the influence of them by smoothing their distribution.

% 2 Related Work
\section{Related Work}
\label{section2}

% 2.1 Cervical Cell Recognition
\subsection{Cervical Cell Recognition}\label{section2.1}
From the analysis of the existing work, extensive research \cite{chankong2014automatic, garcia2016multiscale, sun2017cervical, kudva2018automation, asiedu2018development} has been devoted to the field of automatic recognition of cervical cancer and have obtained good results. The previous cervical cell recognition can be classified into two types according to the number of cells in the image: recognition based on single-cell image and multi-cell image.

Early methods proposed to achieve the automatic segmentation and classification of abnormal cervical cells based on isolated cell  without overlapping  images. The most of them used one or multiple techniques including thresholding \cite{gencctav2012unsupervised}, morphology operation \cite{bamford1996water, anantha2017segmentation}, k-means \cite{tsai2008nucleus}, Hough transform \cite{bergmeir2012segmentation} and watershed \cite{plissiti2010automated}. For the better ways, Li et al. \cite{li2012cytoplasm} utilize a Radiating Gradient Vector Flow (RGVF) Snake to extract both the nucleus and sytoplasm from a single-cell cervical cell image. After preprocessing, the areas in the image are roughly clustered into nucleus, cytoplasm and the background by a spatial K-means clustering algorithm. And then making segmentation of the image by using RGVF after extracting the initial contours. In addition, this work was further improved in \cite{guan2014accurate}, they proposed a dynamic sparse contour searching algorithm to locate the weak contour points of cytoplasm in overlapping regions, and the Gradient Vector Flow Snake model is finally employed to extract the accurate cell contour based on the located contour points. In order to eliminate the limitations of hand-crafted feature, \cite{zhang2017deeppap, yarlagadda2019system, wieslander2017deep} used deep learning for cervical cell classification on single-cell images by extracting and learning features automatically, while the image preprocessing of them rely on effective techniques to deal with the recognition of the nucleus. However, The majority of existing research has done on the Herlve dataset[19] which only contains single-cell images with a size of 200$\times$100 pixels approximately and was produced carefully by trained professionals. Images from Herlve dataset are all clear with no overlapping and impurity, which is different from gigapixel pathological slide in clinical that contains thousands of cervical cells.

Different from recognition based on single-cell image, it is more sophisticated to segment nuclei and cytoplasm of images containing a large number of isolated cells and cell clusters, especially overlapping cells, and this has attracted increasing research interests\cite{song2016accurate, Lu2017Evaluation, lu2015improved, lu2013automated, arya2018clustering}. William et al. \cite{william2019cervical} trained a pixel level classifier on cell nuclei, cytoplasm, background and debris using a Trainable Weka Segmentation(TWS) toolkit \cite{arganda2017trainable} to identify and segment different objects on a slide. Genctav et al. \cite{gencctav2012unsupervised} proposed multi-scale hierarchical segmentation algorithm to partition the image into regions, and used binary classifier to segment cell regions into nucleus and cytoplasm. For better nucleus segmentation, Zhang et al. \cite{zhang2014segmentation} used graph cut approach to segment cervical cells in images with healthy and abnormal cells, and then segmented the nuclei especially abnormal nuclei by using group cut adaptively and locally. In \cite{wang2019automatic}, first obtaining regions of interest for cell nuclei segmentation by applying the Mean-Shift clustering algorithm, and then applying mathematical morphology to split overlapped cell nuclei for better accuracy and robustness.

Although these methods work well for the recognition of cervical cells, it is complicated that all stages should be designed carefully. Furthermore, accurate segmentation of cytoplasm and nucleus for the cervical cell is still particularly challenging due to the complexities of cell structures and image characteristics. Therefore, the recognition of cervical cell based on automatically learning rather than hand-designed is necessary.

% 2.2 CNN-based Object Detection
\subsection{CNN-based Object Detection} \label{section2.2}
In the last years, deep learning is emerging as the leading automatic learning tool in the imaging and computer vision domains generally. Particularly, the CNN-based object detection has the widespread application value and the prospects for development in the field of computer vision tasks. The Overfeat \cite{sermanet2013overfeat} is a pioneer which applies CNN to do object detection and has won the localization task of the ImageNet Large Scale Visual Recognition Challenge 2013(ILSVRC2013)\cite{russakovsky2015imagenet}. Plenty of methods based on CNN have been proposed for object detection consecutively and have achieved significant advances in the last years. The current CNN-based detectors of state-of-the-art can be divided into two categories: (1) the two-stage approach, and (2) the one-stage approach.

In the two-stage approach, a sparse set of candidate object proposal is generated first, and then classify and regress them using convolutional networks. In particular, the famous two-stage approach(e.g., R-CNN \cite{girshick2014rich}, SPPnet \cite{he2015spatial}, Fast R-CNN \cite{girshick2015fast} and Faster R-CNN \cite{ren2015faster}) has been achieving outstanding performances on several challenging benchmarks. Different from the two-stage approach which performs well in accuracy, the one-stage approach detects objects by regular and dense sampling over locations, scale and aspect ratios. The recently one-stage detectors, such as YOLOv3 \cite{redmon2018yolov3}, SSD \cite{liu2016ssd} and the improvements \cite{jeong2017enhancement, fu2017dssd} ,are devoted to high efficiency but most of their accuracy are still yielding to the two-stage detectors.

Recent results have proven that the generic descriptors extracted from CNNs are extremely effective in object recognition and localization in natural images. Extensive research (e.g. \cite{dou2016automatic, teramoto2016automated, liang2018object, bharath2018multi, liang2018end}) has been devoted to the recognition of medical image domain based on deep learning methodologies soon. These methods automatically detect potentially lesion and have been shown to produce impressive and encouraging results. Specifically, CNN-based methodologies have also been used in cervical cell recognition\cite{zhang2017deeppap, wu2018automatic, yarlagadda2019system, wieslander2017deep}, and have achieved good results. However, most of these approaches only used CNNs to classify images with single cell by feeding the network cropped patches they need. In this paper, we treat the cervical cell recognition as object detection and exploit YOLOv3 \cite{redmon2018yolov3} as our cervical cell object detection baseline model, to detect cervical cells directly on multi-cell images and then label the suspicious cells containing location, category, and corresponding confidence score.

%3 Methodology
\section{Methodology}
\label{section3}
\begin{figure}%[htbp]
\centering
\includegraphics[width=0.9\textwidth]{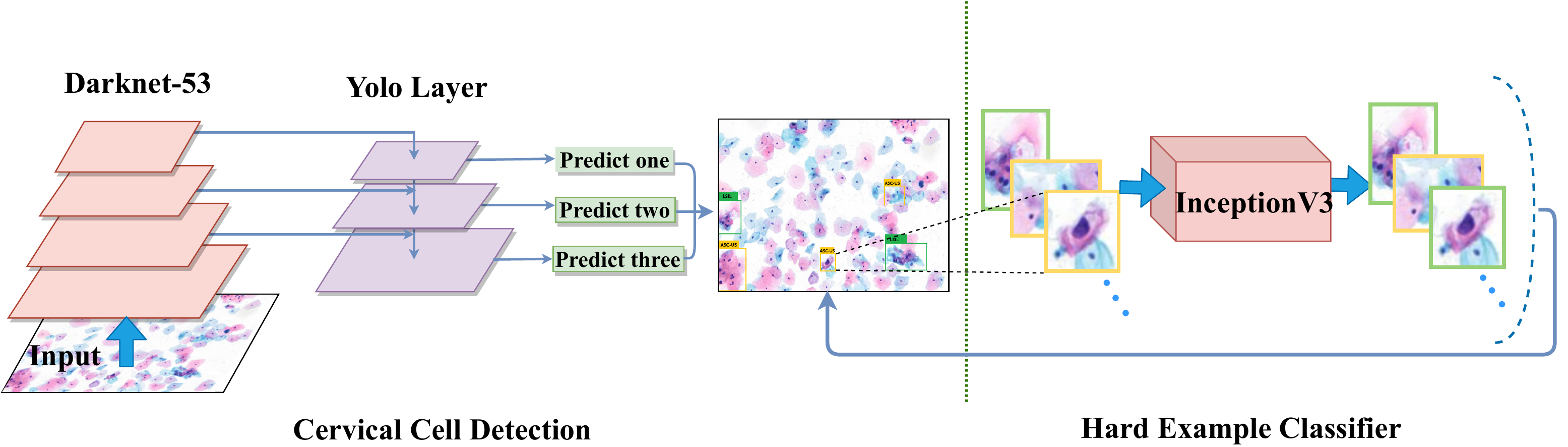}
\caption{\label{figure3}The detail of our proposed methods, cervical cell detection and further cascade the hard example classifier.
}
\end{figure}

The pipeline of proposed method includes cervical cell detection and hard example classifier, as show in Fig.~\ref{figure3}. In this section, we firstly describe the pipeline of proposed end-to-end CNN-based object detection. And then, we introduce several improvements such as hard example classification, smoothing noisy label regularization, to make it more appropriate for our cervical cell recognition.
%figure 3

%3.2.YOLOv3-based cervical cell detector
\subsection{YOLOv3-based Cervical Cell Detector} \label{section3.2}
Most state-of-the-art CNN-based detectors perform well, we accordingly exploit YOLOv3 \cite{redmon2018yolov3} with the input size of 416$\times$416 as our cervical cell object detection framework due to the efficiency, accuracy and flexibility. YOLOv3 \cite{redmon2018yolov3} can be decomposed into two parts, a custom deep architecture Darknet-53, which has 53 layers network trained on ImageNet \cite{deng2009imagenet}, to extract features, and multi-scale feature fusion layers used as feature maps and predictors. In detail, first obtaining feature maps from input image with feature extractor, which is consisted of convolutional layers and residual blocks. Next, predicting on the Yolo layers at three scales, which are given by down-sampling the dimensions of the input image by 32$\times$,16$\times$ and 8$\times$ separately. In fact, small objects are the widespread in our dataset such as vaginalis (abbreviated as VAG), and are difficult to detect well. In order to detect small objects better, the three-scale feature maps fused by up-sampling layers and concatenate layers help preserve the fine-grained features. Finally, we get a 3-d tensor encoding bounding box, objectness score, and class predictions.
%3.3Joint detection and classification
\subsection{Cascade Hard Example Classifier} \label{section3.3}
\begin{figure}%[htbp]
\centering
\includegraphics[width=0.9\textwidth]{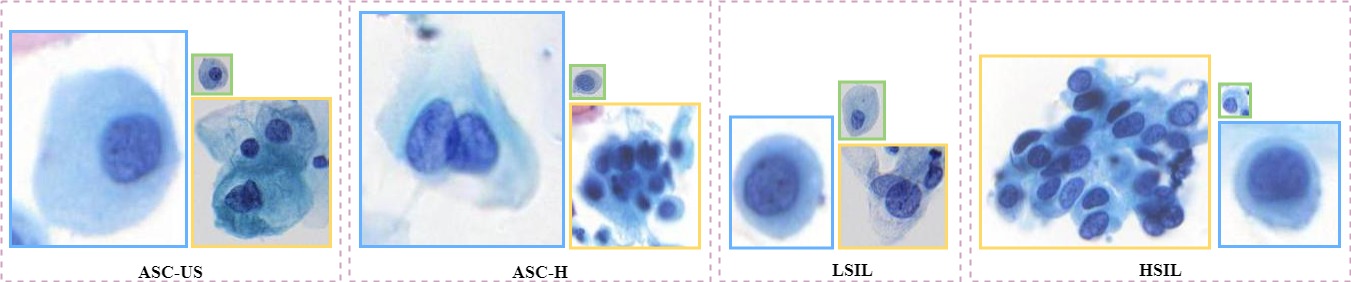}
\caption{\label{figure4}Example images of the hard example objects from our dataset, they are ASC-US, ASC-H, LSIL, HSIL respectively. All these examples keep their originally relative scales after the image pyramid strategy. Note that the blue border means that it is cropped from layer one (4000$\times$3000) of the image pyramid, similarly, yellow border correspond to the layer two (1600$\times$1200), and green border correspond to layer three(800$\times$600).
}
\end{figure}

Cancerization of cells is a continuous process, and many abnormal cells are visually similar to each other. According to section~\ref{section4.1}, the dataset used in this paper contains 10 categories objects from cervical cell images, and the differences between these categories are small, which increasing the difficulty for network to extract more discriminative features and distinguish different cells morphology objects. Especially the four squamous cells categories: atypical squamous cells-undetermined significance (ASC-US), atypical squamous cells-cannot exclude HSIL (ASC-H), low-grade squamous intraepithelial lesion (LSIL), high-grade squamous intraepithelial lesion (HSIL), which are still hard to be distinguished well after object detection networks. As showing in Fig.~\ref{figure4}, differences between these hard example cells are small, and the scales of these cells are diverse after the image pyramid strategy (more details according to section~\ref{section4.1.1}). From the perspective of feature extraction, we can redesign a more fitting backbone network of our dataset to replace Darknet-53. However, it is difficult to obtain a pre-trained model with task-specific identification capability due to the limited dataset and hardware resources. After analysis, cascade a task-specific classifier can also replace the complex backbone network and achieves good performance.

To this end, we implement an object detection and classification cascade framework aiming to improve the hard example identification performance. To be specific, we use InceptionV3 base model \cite{szegedy2016rethinking}, with weights pre-trained on ImageNet \cite{deng2009imagenet}, to achieve good balance between speed and accuracy. On top of the InceptionV3, we attach a Global Average Pooling layer, a fully-connected layer with 1024-dimensional output channels followed by the ReLU function and a fully-connected layer with 4 channels (the number of subclass) followed by the softmax function. In this way, we can obtain more accurate classification of these four classes after the object detection network. As illustrated in the right of Fig.~\ref{figure3}, during testing, for one image which outputted from the cervical cell detector, we just select all bounding boxes which are predicted as the four hard example categories and feed them into cascade classifier to achieve higher accuracy, then feedback the classification results to detection results.
%figure 4
%3.4Smoothing Noisy Label Regularization
\subsection{Smoothing Noisy Label Regularization} \label{section3.4}
Unlike natural images, the medical images are too complex and variable to obtain a perfect annotation as the standard ground truth. In fact, cancerization of cells is a continuous process, there is no absolute threshold for quantitative discrimination that which stage the current cell belongs to. From section~\ref{section3.3} we know that most cells in our dataset are too similar to differentiate them well in most cases whether in clinical or in computer-assisted recognition system at present, especially experts may have different opinions. Therefore, it is extremely tedious and highly subjective to mark hand-crafted labels for cells, which leading to noisy label inevitably.

To cope with this problems, we modify the distribution of our noisy labels to regularize the classifier layer instead of pursuing perfect manual annotation. During training, the input image is computed by the CNNs and gets the confidence score of the current input image corresponding to each category. These scores are normalized by softmax function, and the probability that the current input image belongs to each category is finally obtained:
% equation1
\begin{equation}
p_{i}= \frac{e^{z_{i}}}{ \sum_{j= 1}^{K}e^{z_{j}}} \label{equation1}
\end{equation}
Here, $z_{i}$ are the $logits$ or unnormalized log-probabilities and $K$ is the total number of classes. We set $K$ to 10. And then utilize cross-entropy function to calculate loss:
% equation2
\begin{equation}
Loss= -\sum_{k}^{i=1}q_{i}logp_{i} \label{equation2}
\end{equation}
$q_{i}$ is the ground-truth one-hot distribution of label $i$, where the truth class has probability one while all other classes have zero. Finally, minimizing the cross-entropy of prediction probability and the true probability of label, so as to obtain the optimal prediction probability distribution during training the network. Only when $z_{i}\gg z_{j},\forall j\neq i$
 can softmax function approach this distribution, however, it never reach it. This encourages the model to be too confident about its predictions and leads to over-fitting. Due to the unreliable manual annotation, our model will learn in the wrong direction, for example, the true category of one object is ASC-H, while the manual annotation of it is ASC-US, then our model believe that features extracted from it are belong to ASC-US, but actually not. Label smoothing was proposed in \cite{szegedy2016rethinking} as a form of regularization. Specifically, for a single ground-truth label $i$, we change the probability of true class into $\left ( 1-\varepsilon  \right )$ while all other classes becomes $\varepsilon /K$. We smooth the ground-truth label distribution with
% equation4
\begin{equation}
q_{i}^{'}= \left ( 1-\varepsilon  \right )q_{i}+\frac{\varepsilon }{K} \label{equation4}
\end{equation}
where  $\varepsilon $ is a small constant.  This method can partially reduce the confidence of our model and weakens the influence caused by our noisy labels through suppressing the output difference between positive and negative examples.

% 4.Experiments and Discussion
\section{Experiments and Discussion}
\label{section4}
%4.1.DateSet
\subsection{Image Dataset} \label{section4.1}
%figure 5
\begin{figure}%[htbp]
\centering
\includegraphics[width=0.75\textwidth]{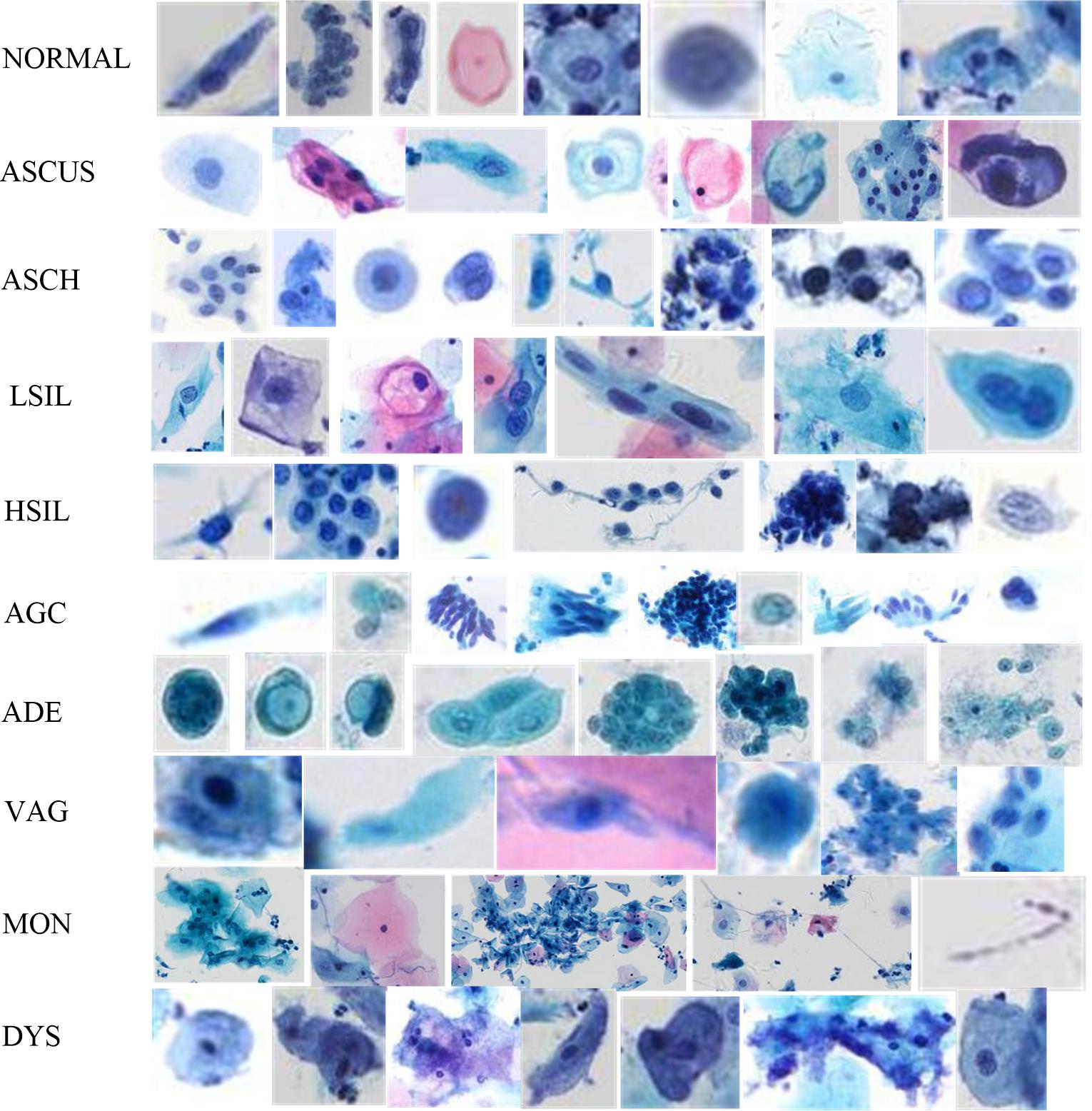}
\caption{\label{figure5}Selected samples of cervical cells from our train set.The same category of cells have multiple features due to  the presence of cell cluster and mucus. Some subsets of different categories have extremely similar features such as the third column of ASCH and the fifth column of HSIL.
}
\end{figure}
As there is no standard clinical cervical cells dataset with multi cells available publicly, we establish our own dataset captured by digital camera Ximea MC124CG-SY-UB with 12 million pixels situated on the microscope Olympus BX40 with 20$\times$ objective. Each pixel has a size of 3.45 $\mu m^{2}$. For one cervical cell slide, we can capture about 1,800 images. The specimens were prepared by liquid-based cytology with Feulgen staining. The dataset used in this paper is consisted of 12,909 cervical images with 58,995 ground truth boxes and contains 10 categories objects from cervical cell images, i.e., normal cells (NORMAL), atypical squamous cells-undetermined significance (ASC-US), atypical squamous cells-cannot exclude HSIL (ASC-H), low-grade squamous intraepithelial lesion (LSIL), high-grade squamous intraepithelial lesion (HSIL), atypical glandular cells (AGC), adenocarcinoma (ADE), vaginalis trichomoniasis (VAG), monilia (MON) and dysbacteriosis (DYS). The ground truth boxes are marked by four trained cyto-technicians and an experienced pathologist, in order to maximize certainty and accuracy of the diagnosis. Examples of 10 categories cervical cells and the visible difference between subcategories are shown in Fig.~\ref{figure5}, we can find that highly similar visual features of inter-class and intra-class are widely exist, such as ASC-H, ASC-US, LSIL and HSIL.
%3.1Data Pre-processing
%\subsection{Data Pre-processing} \label{section4.1}
\subsubsection{Acquisition of Image Patch} \label{section4.1.1}
As the images of our dataset are captured with size of 4112$\times$3008 on the WSI, it is not suitable for the large annotated images to input into our network of which the input size is 416$\times$416, since many deep identification information may lose after compressing and down-sampling in the network. Moreover, enough dataset is crucial to the high performance of CNN. However, as high expertise is required for quality annotation, there is a very limited amount of annotated data for cervical cells available. To deal with these problems, we propose the image pyramid strategy to split the microscopical images into several patches with equally size of 800$\times$600 instead of simple and crude resize operation. In detail, we first resized the original images into three scales,4000$\times$3000 (layer one), 1600$\times$1200 (layer two) and 800$\times$600 (layer three). Then we equally split layer one and layer two into 25 image patches and 4 image patches separately. The size of each patch is 800$\times$600 pixels. Therefore, we got a total of 30 image patches finally. Our image pyramid strategy can expand training samples and obtain multi-scale objects, thereby improving the performance of CNN.

\begin{figure}[htbp]
\centering
\begin{subfigure}[b]{0.23\textwidth}
\includegraphics[width=\textwidth]{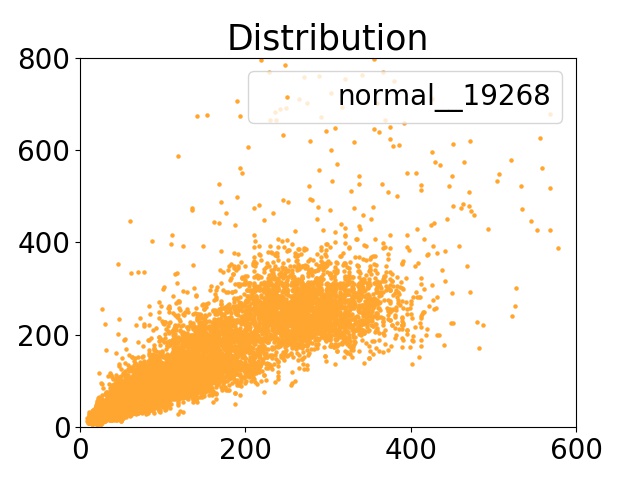}
\caption{NORMAL}
\label{figure4_1}
\end{subfigure}
\begin{subfigure}[b]{0.23\textwidth}
\includegraphics[width=\textwidth]{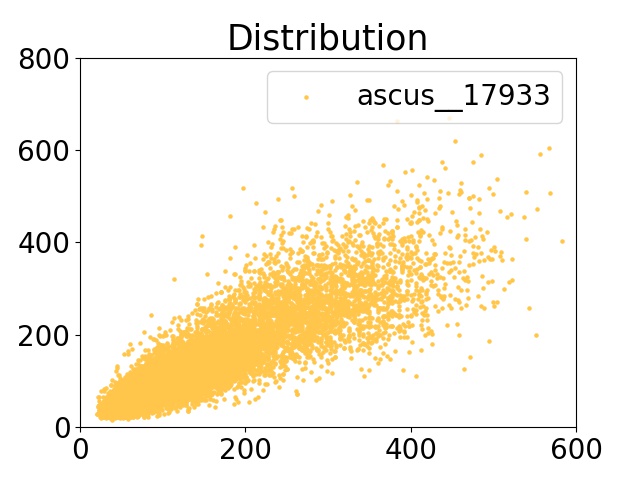}
\caption{ASC-US}
\label{figure4_2}
\end{subfigure}
\begin{subfigure}[b]{0.23\textwidth}
\includegraphics[width=\textwidth]{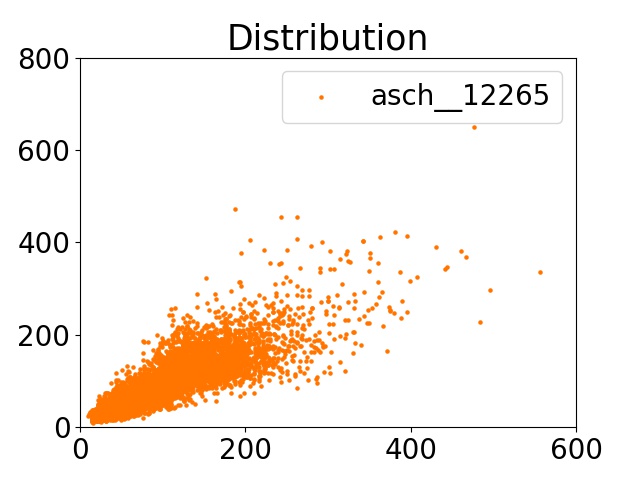}
\caption{ASC-H}
\label{figure4_3}
\end{subfigure}
\begin{subfigure}[b]{0.23\textwidth}
\includegraphics[width=\textwidth]{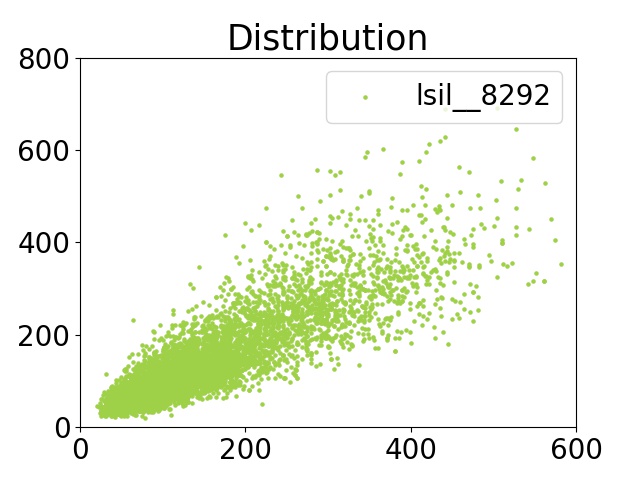}
\caption{LSIL}
\label{figure4_4}
\end{subfigure}
\begin{subfigure}[b]{0.23\textwidth}
\includegraphics[width=\textwidth]{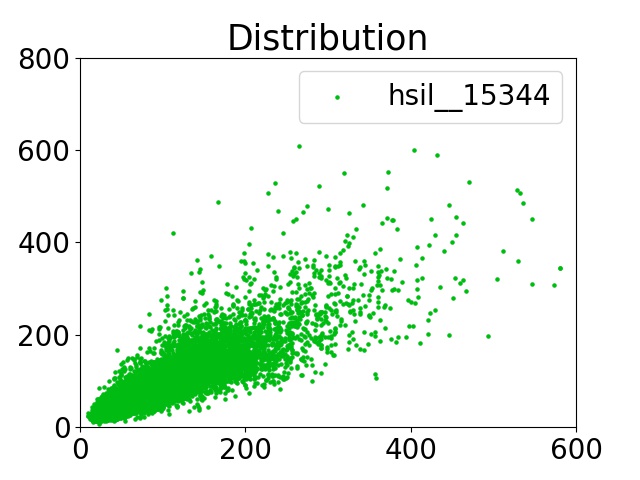}
\caption{HSIL}
\label{figure4_5}
\end{subfigure}
\begin{subfigure}[b]{0.23\textwidth}
\includegraphics[width=\textwidth]{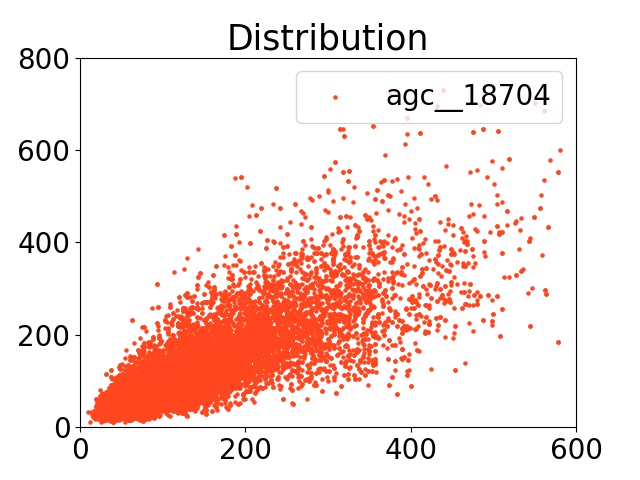}
\caption{AGC}
\label{figure4_6}
\end{subfigure}
\begin{subfigure}[b]{0.23\textwidth}
\includegraphics[width=\textwidth]{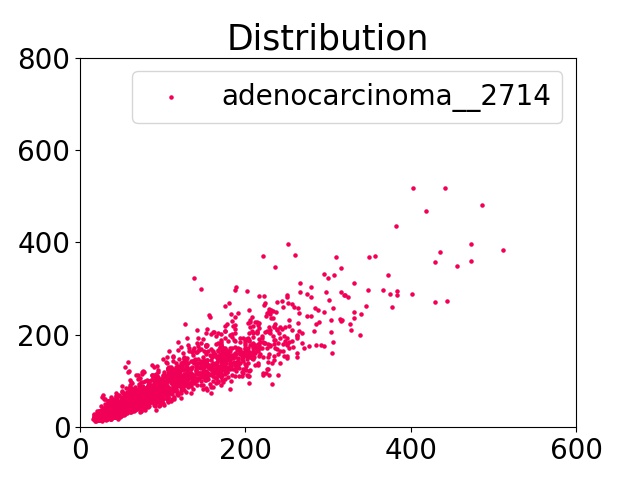}
\caption{ADE}
\label{figure4_7}
\end{subfigure}
\begin{subfigure}[b]{0.23\textwidth}
\includegraphics[width=\textwidth]{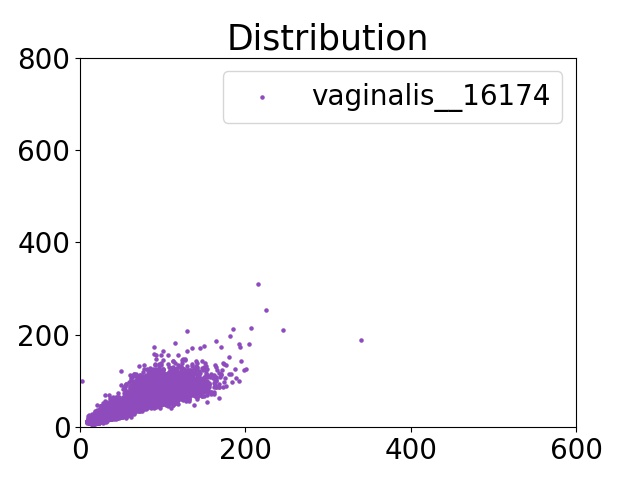}
\caption{VAG}
\label{figure4_8}
\end{subfigure}
\begin{subfigure}[b]{0.23\textwidth}
\includegraphics[width=\textwidth]{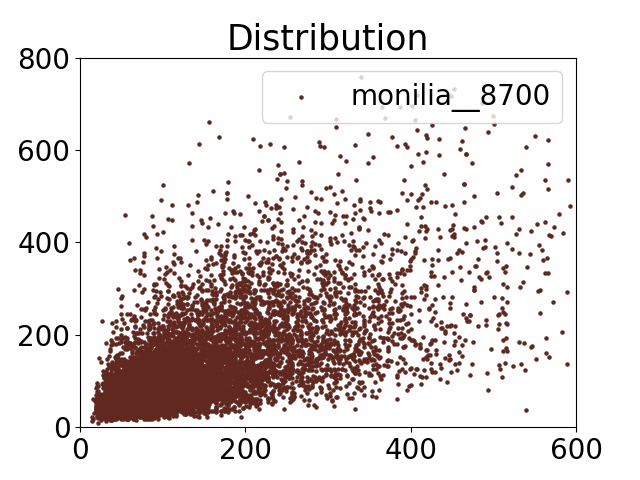}
\caption{MON}
\label{figure4_9}
\end{subfigure}
\begin{subfigure}[b]{0.23\textwidth}
\includegraphics[width=\textwidth]{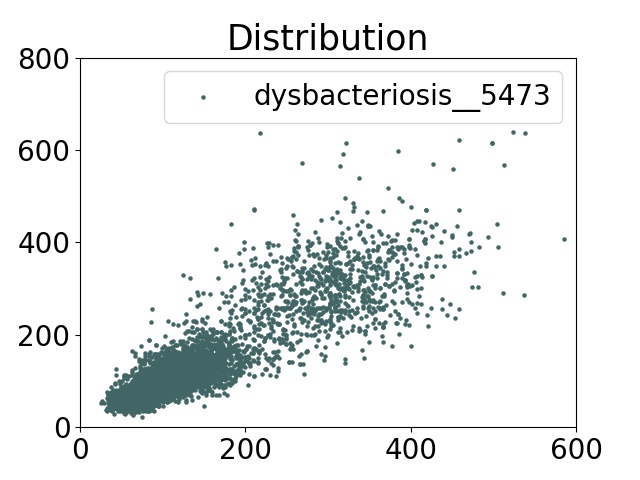}
\caption{DYS}
\label{figure4_10}
\end{subfigure}
\caption{\label{figure6}The size distribution and quantity of each category ground truth boxes after image pyramid strategy. The abscissa axis and ordinate axis indicate the length and width of the instance, respectively. The upper right corner of each graph indicate the total number of instance about each category.
}
\end{figure}

\begin{figure}%[htbp]
\centering
\includegraphics[width=0.5\textwidth]{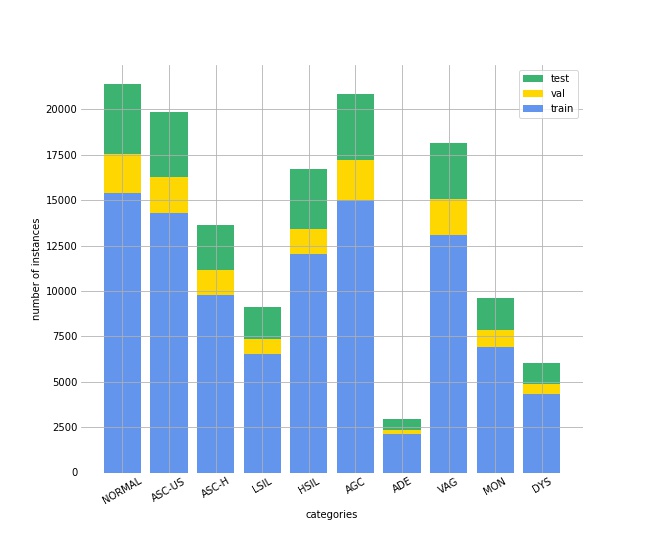}
\caption{\label{figure7}Dataset organization and categories distribution.
}
\end{figure}

As showing in Fig.~\ref{figure6}, the scale of all cells varies widely from 600$\times$800 to 10$\times$10, which brings challenge to our work since smaller objects are difficult to extract richer semantic information and recognize well. After image pyramid strategy, we obtain total 66,627 image patches with size of 800$\times$600 and 138,314 ground truth boxes with multi-scale objects, including 21,388 for NORMAL, 19,879 for ASC-US, 13,616 for ASC-H, 9,092 for LSIL, 16,711 for HSIL, 20,874 for AGC, 2,930 for ADE, 18,173 for VAG, 9,622 for MON, and 6,029 for DYS. From the final datasets, we randomly select 1/10 as test set, and the others as trainval set, where train set makes up 4/5. Fig.~\ref{figure7} demonstrates the details of our dataset organization and categories distribution. The tricolor stacked histogram shows the object quantity of train/ val/ test sets, respectively.
% figure6
%figure 7
%4.2.Evaluation Metrics
\subsection{Evaluation Metrics} \label{section4.2}
As to cervical cell detection evaluation metrics, we follow the evaluation metrics used by the PASCAL VOC object detection challenge \cite{everingham2010pascal}, which are Average Precision (AP) and mean Average Precision (mAP). In this paper, we use AP at a single Intersection over Union (IoU) threshold of 0.5. Since cervical cell detection is a multiple classification task, we calculate AP for detection performance of our model separately for each category. In detail, for a given class, the precision/recall curve is calculated from the ranked ouput of our network. Recall is defined as the proportion of all positive instances ranked above a specific rank. Precision is the proportion of all instances above that rank which are from positive class. The AP summarises the shape of the precision/recall curve, and is defined as the mean precision at a set of eleven equally spaced recall levels[0, 0.1,...,1]\cite{everingham2010pascal}. And mAP is the mean of all categories of AP. In addition, we use accuracy (\emph{Acc}), the global percentage of correctly classified cells, to evaluate the performance of our hard example classifier. As to cervical cell image-level classification, we evaluate the performance of it on our test set with size of 4000$\times$3000 using accuracy (\emph{Acc}), sensitivity (\emph{Sens}) and specificity (\emph{Spec}), where \emph{Sens} means the proportion of correctly classified abnormal images, \emph{Spec} means the proportion of correctly classified normal images, and \emph{Acc} is the total percentage of correctly classified images.

Note that the categories ASC-H and HSIL are too similar to distinguish them confidently in most cases whether in clinical or in computer-assisted recognition system at present, especially experts may have different opinions. To our knowledge, the result that mutual recognition of these two categories is acceptable to the clinical experts. One strategy to cope with the different opinions of experts is to invite some experts to simulate the balance point. We random select 60 cervical cell images which totally contain 52 cells (maybe ASC-H or HSIL), among them 20 objects were predicted as ASC-H (pred-ASCH cells), 34 objects were predicted as HSIL (pred-HSIL cells) by our methods. And then five experts vote the category that each cell belongs to. Finally, for the cells that were predicted as ASC-H, the experts vote 49 ballots to ASC-H and 51 ballots to HSIL, and for the rest cells, the experts vote 112 ballots to ASC-H and 58 ballots to HSIL. In other words, for one cell that was predicted as ASC-H (or HSIL) while the annotated label is HSIL (or ASC-H), we think that our prediction is not absolutely wrong. In order to obtain more reliable and appropriate result for cervical cell recognition, we modify the evaluation metrics by setting TP=0.51 instead of  zero when the detection result is ASC-H but the truth label is HSIL, and correspondingly, setting TP=0.66 instead of zero when the detection result is HSIL while the correct label is ASC-H. Unless otherwise specified, all results are calculated according to the original evaluation metrics.
%4.3.Cervical Cell Recognition Based on YOLOv3
\subsection{Cervical Cell Recognition Based on YOLOv3} \label{section4.3}
We first apply Darknet-53, which originally has 53 layers network trained on Imagenet, as our convolutional feature extractor, to the input image to obtain high-level features. And then fine-tuning all convolutional layers of YOLOV3 on our cervical cell datasets. During training YOLOV3 model, we freeze the first 185 layers of total 252 layers to get a stable loss and the batch size is 32. We started from a learning rate 0.001, and total train 30 epochs. At the time of fine-tuning, we unfreeze all of the layers, with the batch size of 8, and adopt a learning rate policy of plateau: 0.0001 base learning rate, 0.1 factor and patience of 3 (means after 3 epochs while the performance of model does not improve, the learning rate reduction action will be triggered). Generally, we adopt data augmentation, such as rotation, resize,  vertical flip and horizontal flip, in our dataset to expand training samples, avoid over-fitting and improve accuracy. The next series of YOLOv3 experiments are all adopt these strategies and test on 1 NVIDIA GTX1080Ti GPU.

In order to validate the effectiveness of our methods for cervical cell detection, we compared the detection performance of other networks in term of accuracy and speed. Results are reported in Table ~\ref{tab:1}. For the sake of fairness, all these networks did not add any tricks. We find that YOLOv3 takes less test time per image and the mAP is better than the two-stage network FasterR-CNN. We think that the ability of FasterR-CNN to learn multi-scale objects is weak due to the lack of FPN \cite{lin2017feature}structure. We also discussed the input scale and model size of network. Generally, increasing the size of input images can improve the accuracy, so we increase the input size from 416$\times $416 to 608$\times $608. Unfortunately, we just obtain 57.4\%  mAP while cost 99 ms per image.. Tiny YOLOv3 is generated by removing the residual connection \cite{he2016deep} and one yolo layer from YOLOV3. Although the speed is faster than others, the mAP is too worse. We argue that due to the reduction of Yolo layer and simplification of feature extractor, the ability of Tiny YOLOv3 to extract more fine-grained features on multi-scales is decreased.
%table1
\begin{table}[htbp]
\begin{threeparttable}
\caption{\label{tab:1}
Detection accuracy and inference latency with different networks on the our test set, and the image size is 800$\times$600.}
\renewcommand\tabcolsep{1.0pt}
\scriptsize
\centering
%\begin{tabular}{lcccccccccccp{1.2cm}<{\centering}} %表格13列 全部居中显示
\begin{tabular}{p{1.8cm}p{1.2cm}<{\centering}p{1.1cm}<{\centering}
p{1.1cm}<{\centering}p{1.1cm}<{\centering}p{1cm}<{\centering}p{1cm}<{\centering}
p{1cm}<{\centering}p{1cm}<{\centering}p{1cm}<{\centering}p{1cm}<{\centering}
p{1cm}<{\centering}p{1cm}<{\centering}}
\toprule
Method	 & NORMAL & ASCUS & ASCH & LSIL & HSIL & AGC & ADE & VAG & MON & DYS & mAP & Runtime (ms/img)\\
\midrule
FasterR-CNN& 0.432&0.492 &0.340 &0.515 &0.486 &0.787 &0.820 &0.534 &0.547 &0.719 &0.567 &134 \\
YOLOv3 416& 0.606&0.439 &0.360 &0.362 &0.469 &0.794 &0.833 &0.638 &0.482 &0.741 &0.572 &\textbf{65} \\
YOLOv3 608&0.612 &0.439 &0.343 &0.347 &0.467 &0.781 &0.865 &0.638 &0.503 &0.742 &\textbf{0.574} &99 \\
Tiny YOLOv3&0.466 &0.293 &0.300 &0.279 &0.395 &0.655 &0.807 &0.562 &0.249 &0.632 &0.464 &58 \\
\bottomrule
\end{tabular}
\end{threeparttable}
\end{table}

\subsubsection{Re-anchor} \label{section4.3.1}
It is typical to have a collection of boxes overlaid on the image at different spatial locations, scales and aspect ratios that act as "anchors" (also called "priors" or "default boxes") \cite{huang2017speed}. Generally, YOLOv3 \cite{redmon2018yolov3} used k-means clustering on the COCO dataset \cite{lin2014microsoft} to determine their bounding box priors as anchor boxes. Unlike generic objects in natural images, the objects of cervical cells vary very widely in their shapes, sizes and numbers, which lead to poor location and regression performance of potential instances. Therefore, we used k-means clustering to generate 9 anchors based on our dataset again (called "re-anchor" by the authors). Our new 9 clusters were: (7 $\times$ 11), (12 $\times$ 19), (17 $\times$ 26), (26 $\times$ 36), (32 $\times$ 52), (47 $\times$ 64), (60 $\times$ 94), (92 $\times$ 127), (144 $\times$ 208). Table ~\ref{tab:3} shows that our new anchors are more adaptable and robust in our dataset than the original anchor that generated based on COCO, and yields higher mAP.
%4.4.Ablation Studies
\subsection{Ablation Studies} \label{section4.4}
For all ablation studies, we use an image scale of 800$\times$600 for both training and testing. We evaluate the contribution of several important elements to our methods, including hard example classification, smoothing noisy label regularization, and re-anchor. The performance of our methods are all calculated on test specimens with size of 4000$\times$3000. Results as reported in Table ~\ref{tab:3}.
%table3
\begin{table}[htbp]
\begin{threeparttable}
\caption{\label{tab:3}
Ablative experiments for our methods. YOLOv3 is the base network for all experiments in this table. We study the contribution of hard example classification, smoothing noisy label regularization and re-anchor. The last row denotes the final results using the improved evaluation metrics, and bold numbers mean the results of ASC-H and HSIL. “HEC” denotes that using hard example classification. “L-S” denotes that using smoothing noisy label regularization. “R-A” denotes that using re-anchor strategy. }
\renewcommand\tabcolsep{1.0pt}
\scriptsize
\centering
%\begin{tabular}{lcccccccccccp{1.2cm}<{\centering}} %表格13列 全部居中显示
\begin{tabular}{p{1.1cm}<{\centering}p{1.1cm}<{\centering}p{1.1cm}<{\centering}p{1.2cm}<{\centering}
p{1.1cm}<{\centering}p{1.1cm}<{\centering}p{1cm}<{\centering}p{1cm}<{\centering}
p{1cm}<{\centering}p{1cm}<{\centering}p{1cm}<{\centering}p{1cm}<{\centering}
p{1cm}<{\centering}p{1cm}<{\centering}}
\toprule
HEC & L-S & R-A	& NORMAL & ASCUS & ASCH & LSIL & HSIL & AGC & ADE & VAG & MON & DYS & mAP \\
\midrule
& & & 0.606&0.439 &0.360 &0.362 &0.469 &0.794 &0.833 &0.638 &0.482 &0.741 &0.572 \\
& & $\surd$& 0.622&0.459 &0.365 &0.393 &0.481 &0.802 &0.797 &0.692 &0.517 &0.750 &0.588 \\
$\surd$& & & 0.606&0.425 &0.419 &0.507 &0.443 &0.794 &0.833 &0.638 &0.482 &0.741 &0.589 \\
& $\surd$& &0.630 &0.458 &0.363 &0.388 &0.467 &0.815 &0.821 &0.655 &0.505 &0.741 &0.584 \\
& $\surd$& $\surd$&0.630 &0.469 &0.369 &0.361 &0.459 &0.801 &0.855 &0.704 &0.504 &0.765 &0.592 \\
$\surd$& $\surd$& &0.630 &0.438 &0.403 &0.515 &0.432 &0.816 &0.821 &0.655 &0.505 &0.741 &0.596 \\
$\surd$& & $\surd$&0.622 &0.437 &0.378 &0.425 &0.508 &0.802 &0.797 &0.692 &0.517 &0.750 &0.593 \\
$\surd$& $\surd$ &$\surd$&0.630 &0.437 &0.399 &0.500 &0.421 &0.801 &0.855 &0.704 &0.504 &0.765 &0.602 \\
\midrule
$\surd$& $\surd$ &$\surd$&0.630 &0.437 &\textbf{0.627} &0.500 &\textbf{0.519} &0.801 &0.855 &0.704 &0.504 &0.765 &0.634 \\
\bottomrule
\end{tabular}
\end{threeparttable}
\end{table}

\subsubsection{Hard Example Classification} \label{section4.4.1}
Same as detection, we utilize InceptionV3 base model \cite{szegedy2016rethinking}, with weights pre-trained on ImageNet \cite{deng2009imagenet}, to achieve good balance between speed and accuracy. During training, we fine-tune all Inception modules by freezing the first 17 layers of total 314 layers after attaching our classification layers and the batch size is 32. And also adopting a learning rate policy of plateau: 0.0001 base learning rate, 0.1 factor and patience of 5 epochs. The input size of the network is 299 $\times$ 299. As there are no original classification datasets of cervical cell, we produce the classification datasets by cropping the original annotated detection images according to their ground truth bounding box. In fact, our classification dataset is consisted of 16,332 squamous cell images as train set, 2,082 squamous cell images as validation set and 2447 as test set, which contain 4 categories, ASC-US, ASC-H, LSIL, HSIL.  Commonly, we also adopt data augmentation, such as rotation, shift, shear, zoom, and random horizontal flip, in our train set.
%table4
\begin{table}[htbp]
\caption{\label{tab:4}
The comparison of classification performance using different pre-trained model.}
\renewcommand\tabcolsep{1.0pt}
\scriptsize
\centering
%\begin{tabular}{lcccccccccccp{1.2cm}<{\centering}} %表格13列 全部居中显示
\begin{tabular}{p{2.2cm}p{1.8cm}<{\centering}p{1.8cm}<{\centering}}
\toprule
Model & Model Size & Acc \\
\midrule
VGG19&549MB &0.667 \\
MobileNet&17MB &0.691 \\
Xception&88MB &0.713 \\
InceptionV3&92MB &0.701 \\
\bottomrule
\end{tabular}
\end{table}

As expected, simply cascade the object detection and hard example classification greatly improves the mAP by 1.7\%, especially ASC-H and LSIL on which improve the AP by 5.9\% and 14.5\% respectively, shown in Table ~\ref{tab:3}. The further task-specific classifier can extract and learn more discriminative features on these hard examples. What’s more, we find that the hard example classification after detection networks with our new anchor boxes can improve the mAP by 2.1\% (the seventh row of Table ~\ref{tab:3}). As our new anchor boxes are more adaptable and robust in our dataset, they are beneficial to the location accuracy and accelerate convergence. Similarly, we also make experiments about varies classification model aiming to select the optimal model that is appropriate to our dataset. We fine-tune Xception \cite{chollet2017xception}, MobileNet \cite{howard2017mobilenets} and VGG19 \cite{simonyan2014very} as a base model to train our classification and adopted the same training strategy with InceptionV3. Table~\ref{tab:4} presents the comparison, from which we can see that VGG19 performs poor than InceptionV3, the model size is 6$\times$bigger. Although the Acc of Xception \cite{chollet2017xception} is better, the training time is double than InceptionV3 based on the similar model size as decreasing the batch size setting from 32 to 16 to run Xception \cite{chollet2017xception} in limited GPU resources.

\subsubsection{Smoothing Noisy Label Regularization} \label{section4.4.2}
Instead of pursuing perfect manual annotation, we can modify the distribution of our noisy labels to regularize the classifier layer. We smooth the ground-truth label distribution with Eqn.~\ref{equation4} where $\varepsilon$ is a small constant. In order to achieve higher incremental in our dataset, we make several experiments with different $\varepsilon$.  The comparative results are listed in Table ~\ref{tab:5}, which shows that $\varepsilon = 0.1$ yield higher mAP in general.
As stated in Table ~\ref{tab:3}, smoothing noisy label regularization performs well on our dataset. It improves the mAP by 1.2\% and 0.7\% on the baseline and our network cascade with hard example classifier respectively. Finally, with re-anchor, the mAP is further improve to 60.2\%. We make the observation that improvement mainly comes from the more accurate classification on hard example and more accurate localization. We also find that smoothing noisy label regularization and re-anchor can work well together. Applying smoothing noisy label regularization with new anchor boxes can improve the mAP by 2.0\%. We argue that clustering on our dataset makes the network selects more appropriate priors anchor boxes which can improve the location accuracy and accelerate convergence. And modify the distribution of our noisy labels and further cascade classifer can improve the classification performance. Note that the AP of ASC-H and HSIL improves by 22.8\% and 9.8\% respectively using our improved evaluation metrics (the last row in Table ~\ref{tab:3}). We demonstrate that mutual recognition between them is ordinary.
%table5
\begin{table}[htbp]
\caption{\label{tab:5}
Comparison of detection performance with different $\varepsilon $.}
\renewcommand\tabcolsep{1.0pt}
\scriptsize
\centering
\begin{tabular}{p{1.5cm}p{1.5cm}<{\centering}p{1.5cm}<{\centering}p{1.5cm}<{\centering}
p{1.5cm}<{\centering}p{1.5cm}<{\centering}p{1.5cm}<{\centering}p{1.5cm}<{\centering}p{1.5cm}<{\centering}p{1.5cm}<{\centering}}
\toprule
$\varepsilon $ &0.01 & 0.05 &0.10 &0.11 &0.12 &0.15 &0.20 &0.30 &0.50 \\
\midrule
mAP&0.557 & 0.547 &0.583 &0.552 &0.568 &0.567 &0.566 &0.529 &0.542 \\
\bottomrule
\end{tabular}
\end{table}
% classification on cervical cell specimens
\subsection{Classification on Cervical Cell Images} \label{section4.5}
In order to evaluate the image-level classification performance of our method on cervical cell dataset, we calculate the \emph{Acc}, \emph{Sens}, and \emph{Spec} evaluation metrics on test set. According to the Bethesda system (TBS) for reporting cervical cytology \cite{nayar2015bethesda}, precancerous lesions and cancerous lesions as positive samples in our test set include five types: ASC-US, LSIL, ASC-H, HSIL, AGC and ADE. In detail, our test set contains total 1,014 cervical cell images with size of 4000$\times$3000, which are consisted of 728 abnormal cell images (positive samples) and 286 normal cell images (negative samples). We test both on our baseline model and totally improved method above. From Table ~\ref{tab:6} we can see that the accuracy and sensitivity are high both on our baseline model and improved method, which illustrate the feasibility of our idea on image-level classification. In detail, we achieve a \emph{Acc} with 89.3\% and a perfect \emph{Sens} with 97.5\%, which is higher than most of previous approaches. Moreover, different from most of the traditional approaches operated on single-cell images, the primary screening results given by us is calculated on multi-cell images with size of 4000$\times$3000. In other words, we can provide image-level assisted reference information to cytotechnologists and doctors instead of cell-level, which boosts efficiency of cervical cell primary screening. We make the observation that the \emph{Spec} value is low with 67.8\%. We consider such poor results mainly from the extremely imbalanced data distribution (number of positive samples is 3$\times$  than negative samples), which induces the model to identified more cells as abnormal. However, in the practical application for automation-assisted primary screening in clinical, high \emph{Sens} even with fairly low \emph{Spec} is acceptable for image-level screening due to all positive samples will be reexamined by doctors or cyto-experts. Furthermore, our improved methods substantially decreases \emph{FP} and increases \emph{Spec} by 5.6\%. Therefore, we believe that the performance of our method on image-level classification is satisfactory and inspiring.

%table6
\begin{table}[htbp]
\caption{\label{tab:6}
Classification performance of our proposed method on total 1,014 test images with size of 4000$\times$3000. \emph{TP} represents correctly classified image as the abnormal image, \emph{FP} represents incorrectly classified normal image as abnormal image, \emph{TN} represents correctly classified image as normal image, and \emph{FN} represents incorrectly classified abnormal image as the normal image.}
\renewcommand\tabcolsep{1.0pt}
\scriptsize
\centering
\begin{tabular}{p{2.2cm}p{0.9cm}<{\centering}p{0.9cm}<{\centering}p{0.9cm}<{\centering}p{0.9cm}<{\centering}p{1.1cm}<{\centering}p{1.1cm}<{\centering}p{1.1cm}<{\centering}}
\toprule
Method &\emph{TP} &\emph{FP} &\emph{TN} &\emph{FN} &\emph{Acc}(\%)&\emph{Sens}(\%)&\emph{Spec}(\%) \\
\midrule
Baseline &716 & 109 &178 &12 &88.1 &\textbf{98.4} &62.2 \\
Improved method&710 & 92 &194 &18 &\textbf{89.3} &97.5 &67.8\\
\bottomrule
\end{tabular}
\end{table}
% 5.Conclusions
\section{Conclusions}
\label{section5}
In this paper, we utilize object detection method to achieve the automation-assisted cervical cell reading system. Different from the multi-stage traditional approaches, which rely on the accuracy of segmentation and the efficiency of hand-crafted features, our method extract high-level features automatically and detect cervical cells directly. We exploit YOLOv3 as a base model to detect 10 categories and then cascaded a further hard example classifier to refine the 4 categories: ASC-US, ASC-H, LSIL, HSIL. Finally, we also investigated the presence of the noisy label and deal with them by smoothing their distribution. After conducting comprehensive experiments, the image-level classification performance on cervical cell test set with size of 4000$\times$3000 is excellent with 89.3\% \emph{Acc},97.5\% \emph{Sens} and 67.8\% \emph{Spec}. Particularly, not only dose our method achieve the automatic cervical cell image-level classification screening but also it can detect the categories and location of abnormal cells. We achieve a mean average precision(mAP) of 60.2\%  in our dataset, and improve the AP of hard examples which are the most valuable but most difficult to distinguish. The results indicate that the performance of our automatic detection method provides a good reference and basis for the next work. We hope that our methods can provide a new idea for future development of computer-assisted reading systems for cervical cell primary screening and diagnosis in clinical.
\section*{Conflict of interest}
\addcontentsline{toc}{section}{Conflict of interest}
The authors declare that there is no conflict of interest regarding the publication of this paper.
% 6Acknowledgements
\section*{Acknowledgements}
\addcontentsline{toc}{section}{Acknowledgements}
This work was partially supported by the National Natural Science Foundation of China under Grant No.61602522 , and the Fundamental Research Funds of the Central Universities of Central South University [No.2018zzts595].

\clearpage
\section*{Appendix}
\addcontentsline{toc}{section}{Appendix}
\begin{figure}[ht]
\centering
\begin{subfigure}[b]{1.0\textwidth}
\includegraphics[width=\textwidth]{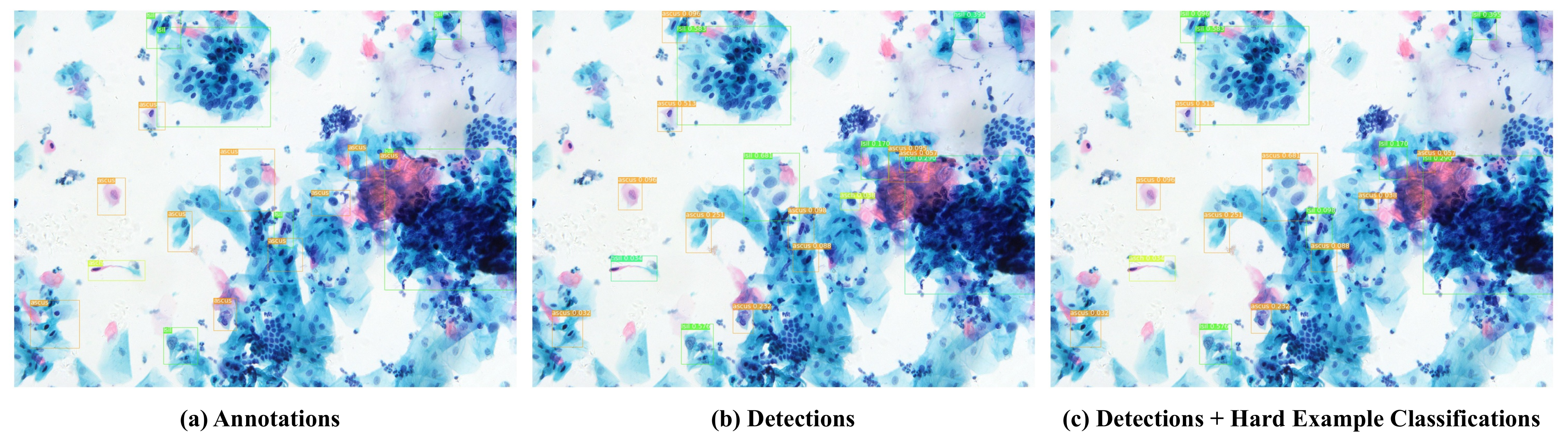}
\caption{Detection and classification results of hard examples. }
\label{s1}
\includegraphics[width=\textwidth]{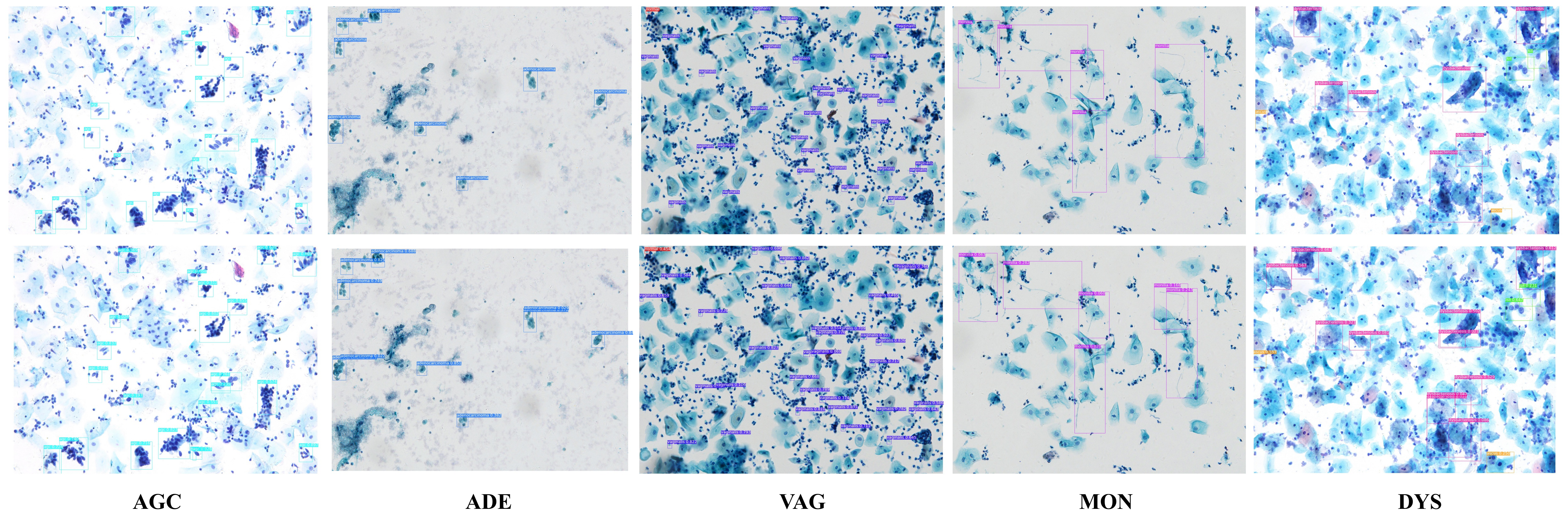}
\caption{Detection results of other abnormal categories. First row: the specimen and ground truth boxes. Bottom row: Corresponding detection results. }
\label{s2}
\end{subfigure}
\caption{\label{figure8}Selected detection specimens (with size of 4000$\times$3000) of abnormal cervical cells on test set. We show results with scores which are calculated by multiplying classification scores and location scores. All abnormal categories are divided into two groups, where one group are hard example categories and another group are gland cells and microbial cells categories. In each group, \textbf{a} shows original image with ground truth boxes, \textbf{b} are detection results on YOLOv3 with new anchor boxes, \textbf{c} shows the results after cascading the detector and hard example classifier. For the ground truth boxes and detection boxes, different categories use only different colors: normal (red), ascus (orange), asch (yellow), lsil (fluorescent green), hsil (green), agc(lake blue), ade (blue), vag (purple), mon (rose red), dys (pink). As shown in this figure, the performance of our methods are acceptable, and the hard example classifier is effective and necessary. Viewing digitally with zoom is recommended. }
\end{figure}

\addcontentsline{toc}{section}{References}
\bibliography{paper_TCT}

\end{document}